\documentclass{article}

\usepackage{graphicx} 
\usepackage{subfigure} 

\usepackage{natbib}

\usepackage{algorithm}
\usepackage{algorithmic}
\usepackage{latexsym}
\usepackage{amsmath}
\usepackage{amsthm}
\usepackage{amsfonts}
\usepackage{graphicx}
\usepackage{url}
\usepackage{hyperref}
\usepackage{float}

\newcommand{\cond}{\mathbin{|}}
\newcommand{\ind}{\perp\!\!\!\perp}

\newtheorem{definition}{Definition}
\newtheorem{theorem}{Theorem}

\newtheorem{corollary}{Corollary}

\usepackage[accepted]{icml2013}

\icmltitlerunning{Separate Training for Conditional Random Fields}

\begin{document} 

\twocolumn[
\icmltitle{Separate Training for Conditional Random Fields Using \\ Co-occurrence Rate Factorization \\(Technical Report: TR-CTIT-12-29)}

\icmlauthor{Zhemin Zhu}{Z.Zhu@utwente.nl}
\icmlauthor{Djoerd Hiemstra}{D.Hiemstra@utwente.nl}
\icmlauthor{Peter Apers}{P.M.G.Apers@utwente.nl}
\icmlauthor{Andreas Wombacher}{a.wombacher@utwente.nl}
\icmladdress{PO Box 217, CTIT Database Group, University of Twente, Enschede, the Netherlands}

\icmlkeywords{Conditional Random Fields, Graphical Models}

\vskip 0.3in
]

\begin{abstract} 
The standard training method of Conditional Random Fields (CRFs) is very slow for large-scale applications. As an alternative, piecewise training divides the full graph into pieces, trains them independently, and combines the learned weights at test time. In this paper, we present \emph{separate} training for undirected models based on the novel Co-occurrence Rate Factorization (CR-F). Separate training is a local training method. In contrast to MEMMs, separate training is unaffected by the label bias problem. Experiments show that separate training (i) is unaffected by the label bias problem; (ii) reduces the training time from weeks to seconds; and (iii) obtains competitive results to the standard and piecewise training on linear-chain CRFs.
\end{abstract}

\section{Introduction}
Conditional Random Fields (CRFs) \cite{Lafferty:2001:CRF:645530.655813} are undirected graphical models that model conditional probabilities rather than joint probabilities. Thus CRFs do not need to assume the unwarranted independence over observed variables. CRFs define a distribution conditioned by the whole observed variables. This global conditioning allows the use of rich features, such as overlapping and global features. CRFs have been successfully applied to many tasks in natural language processing \cite{McCallum:2003:ERN:1119176.1119206,  Sha:2003:SPC:1073445.1073473, Cohn:2005:SRL:1706543.1706573, Blunsom:2006:DWA:1220175.1220184} and other areas.

Despite the apparent successes, the training of CRFs can be very slow \cite{Sutton05piecewisetraining, Cohn07scalingconditional, sutton_intro}. In the standard training method, the calculation of the global partition function $Z(X)$ is expensive \cite{Sutton05piecewisetraining}. The partition function depends not only on model parameters but also on the input data. When we calculate the gradients and $Z(X)$ using the forward-backward algorithm, the intermediate results can be efficiently reused by dynamic programming within a training instance, but they can not be reused between different instances. Thus we have to calculate $Z(X)$ from scratch for each instance in each iteration of the numerical optimization. On linear-chain CRFs, the time complexity of the standard training method is quadratic in the size of the label set, linear in the number of features and almost quadratic in the size of the training sample \citep{Cohn07scalingconditional}. In our POS tagging experiment (Tab. \ref{tab:32623-1000}), the standard training time is up to several weeks even though the graph is a simple linear chain. Slow training prevents applying CRFs to large-scale applications.

To speed up the training of CRFs, \emph{piecewise} training \cite{Sutton05piecewisetraining} decomposes the full graph into pieces, trains them independently and combines the learned weights in decoding. At training time, \emph{piecewise} training replaces the exact global partition function in the maximum likelihood objective function with its upper bound approximation. This upper bound approximation is the summation of the partition functions restricted on disjoint pieces such as edges. The pieces can be generalized from edges to factors of higher arity. Whenever the pieces are tractable, the partition functions restricted on pieces can be calculated efficiently. The upper bound of piecewise training is derived from tree-reweighed upper bound \cite{Wainwright:2002:NCU:2073876.2073940}, where the upper bound of the exact global log partition function is  a linear combination of the partition functions restricted on tractable subgraphs such as spanning trees. An unsolved problem in piecewise training is what is a good choice of pieces. There is a similar problem in the choice of regions of generalized belief propagation (GBP) \cite{Yedidia00generalizedbelief}. \citet{Welling:2004:CRG:1036843.1036914} gives a solution using operations (Split, Merge and Death) on region graphs which leave the free energy and sometimes the fixed points of GBP invariant. Experiment results show piecewise training obtains better results in two of the three NLP tasks than the standard training. As BP on linear-chain graphs is exact, it is a surprise that approximate training may outperform exact training. After personal communication, \citet{Cohn07scalingconditional} attributes this to the exact training over-fitting the data. The piecewise training may smooth the over-fitting model to some degree. This also happens in Maximum Entropy Markov Models (MEMMs) \cite{Mccallum00maximumentropy} which factorize the joint distribution into small factors. But MEMMs suffer from the label bias problem \cite{Lafferty:2001:CRF:645530.655813} which offsets this smoothing effect.

In this paper, we present \emph{separate} training. Separate training is based on the Co-occurrence Rate Factorization (CR-F) which is a novel factorization method for undirected models. In separate training, we first factorize the full graph into small factors using the operations of CR-F. This also means the selection of factors (pieces) is not as flexible as piecewise training. Then these factors are trained separately. In contrast to directed models such as MEMMs, separate training is unaffected by the label bias problem. Experiment results show separate training performs comparably to the standard and piecewise training while reduces training time radically.

\section{Co-occurrence Rate Factorization}
\label{sec_CR}
Co-occurrence Rate (CR) factorization is based on elementary probability theory. CR is the exponential function of Pointwise Mutual Information (PMI) \cite{fano1961} which was first introduced to NLP community by \citet{Church:1990:WAN:89086.89095}. PMI instantiates Mutual Information \cite{Shannon1948} to specific events and was originally defined between two variables. To our knowledge, the present work is the first to apply this concept to factorize \emph{undirected} graphical models in a systematic way. 

\textbf{Notations} A graphical model is denoted by $G=(X_G, E_G)$, where $X_G=\{X_1,...,X_{\mathbin{\cond}X_G\mathbin{\cond}}\}$ are nodes denoting random variables, and $E_G$ are edges. The joint probability of the random variables in $X_A$, where $X_A\subseteq X_G$, is denoted by $P(X_A)$. $X_{\emptyset}$ is the empty set of random variables.
\begin{definition}[Discrete CR]
\label{def_CR_dis}
Co-occurrence rate between discrete random variables is defined as:
\vspace*{-3mm}
\small
\begin{align*}
&CR(X_1;...;X_n)= \frac{P(X_1,...,X_n)}{P(X_1)...P(X_n)},\,\,\, \text{\textbf{if}}\,\, n\geq 1\\
&CR(X_{\emptyset})=1,
\end{align*}
\normalsize
where $X_1,...,X_n$ are discrete random variables, and $P$ is probability.
\end{definition}
Singleton CRs which contain only one random variable are equal to $1$. In Thm. (\ref{thm_marginal_cr}) we will explain the reason to define $CR(X_{\emptyset})=1$. If any singleton marginal probabilities in the denominator equals 0, then CR is undefined. $CR$ is a non-negative quantity with clear intuitive interpretation: (i) If $0\leq CR<1$, events occur \emph{repulsively}; (ii) If $CR=1$, events occur \emph{independently}; (iii) If $CR>1$, events occur \emph{attractively}. 

We distinguish the following two notations:
\vspace*{-1mm}
\small
\begin{align*}
&CR(X_1;X_2;X_3)=\frac{P(X_1,X_2,X_3)}{P(X_1)P(X_2)P(X_3)},\\
&CR(X_1;X_2X_3)=\frac{P(X_1,X_2,X_3)}{P(X_1)P(X_2,X_3)}.
\end{align*}
\normalsize
The first one denotes $CR$ between three random variables: $X_1$, $X_2$ and $X_3$. By contrast, the second one denotes $CR$ between two random variables:  $X_1$ and the other joint random variable $X_2X_3$. We will use the following two different notations to distinguish them explicitly when we manipulate a set of variables:
\begin{align*}
&Sem\,X_A:=X_1;X_2;...;X_n\\
&Seq\,X_A:=X_1X_2...X_n
\end{align*}
$Sem$ and $Seq$ stand for $Semicolon$ and $Sequence$, respectively.

\begin{definition}[Continuous CR]
\label{def_CR_con}
Co-occurrence rate between continuous random variables is defined as:
\begin{align*}
&CR(X_1;...;X_n)= \frac{p(X_1,...,X_n)}{p(X_1)...p(X_n)},
\end{align*}
where $n\geq 1$, $X_1,...,X_n$ are continuous random variables, and $p$ is the probability density function. 
\end{definition}
Continuous CR preserves the same semantics as the discrete CR:

\vspace{-3mm}
\small
\begin{align}
\nonumber &CR(X_1;...;X_n)\\
\nonumber &=\lim_{\varepsilon \downarrow 0}\frac{P(x_1 \leq X_1\leq x_1+
\epsilon_1 ,..., x_n \leq X_n\leq x_n + \epsilon_n )}{P(x_1 \leq X_1\leq x_1+
\epsilon_1 )...P(x_n \leq X_n\leq x_n + \epsilon_n )}\\
\nonumber &=\lim_{\varepsilon \downarrow 0}\frac{\int_{x_1}^{x_1+
\epsilon_1}...\int_{x_n}^{x_n+ \epsilon_n}p(X_1,...,X_n)dX_1...dX_n}{\int_{x_1}^{x_1+
\epsilon_1}p(X_1)dX_1...\int_{x_n}^{x_n+\epsilon_n}p(X_n)dX_n}\\
\nonumber &=\lim_{\varepsilon \downarrow 0}\frac{\epsilon_1 ...\epsilon_n p(X_1,...,X_n)}{\epsilon_1 p(X_1)... \epsilon_n p(X_n)}=\frac{p(X_1,...,X_n)}{p(X_1)...p(X_n)},
\end{align}
\normalsize
where $\varepsilon=\{\epsilon_1,...,\epsilon_n\}$. In the rest of this paper, we only discuss the discrete situation. The results can be extended to the continuous case.

\begin{definition}[Conditional CR]
\label{def_conditional_cr}
The Co-occurrence rate between $X_1,...,X_n$ conditioned by $Y$ is defined as:
\begin{equation*}
CR(X_1;...;X_n\mathbin{\cond}Y)=\frac{P(X_1,...,X_n\mathbin{\cond}Y)}{P(X_1\mathbin{\cond}Y)...P(X_n\mathbin{\cond}Y)}.
\end{equation*}
\end{definition}
In the rest of this section, the theorems which are given in the form of unconditional CR also apply to Conditional CR, which can be easily proved.

The joint probability and conditional probability can be rewritten by CR:

\vspace{-3mm}
\small
\begin{align}
\label{eqn_joint_CR}
&P(X_1,...,X_n)=CR(X_1;...;X_n)\prod_{i=1}^n P(X_i)\\
\nonumber
&P(X_1,...,X_n\mathbin{\cond}Y)=CR(X_1;...;X_n | Y)\prod_{i=1}^n P(X_i|Y).
\end{align}
\normalsize
Instead of factorizing the joint or conditional probability on the left side, we can first factorize the joint or conditional $CR$ on the right side.

\begin{theorem}[Conditioning Operation] 
\label{thm_cond_op}
\small
\begin{align*}
&CR(X_1;...;X_n)=\frac{P(X_1,...,X_n)}{P(X_1)...P(X_n)}=\frac{\sum_Y P(X_1,...,X_n,Y)}{P(X_1)...P(X_n)}\\
&=\frac{\sum_Y CR(X_1;...;X_n; Y)P(X_1)...P(X_n)P(Y)}{P(X_1)...P(X_n)}\\
&=\sum_Y CR(X_1;...;X_n\cond Y)CR(X_1;Y)...CR(X_n;Y)P(Y).
\end{align*}
\end{theorem}
\normalsize
This theorem builds the relation between $CR(X_1;...;X_n)$ and $CR(X_1;...;X_n\cond Y)$, and can be used to break loops (Sec. \ref{sec_fu}).

\begin{theorem}[Marginal CR] Let $n\geq 1$,
\label{thm_marginal_cr}

\vspace{-3mm}
\small
\begin{equation*}
\sum_{X_n}[CR(X_1;...;X_{n-1}; X_n)P(X_n)]= CR(X_1;...;X_{n-1}).
\end{equation*}
\normalsize
\end{theorem}
This theorem allows to reduce random variables existing in $CR$. If we want this theorem still hold when $n=1$, we need to define $CR(X_{\emptyset})=1$ (Def. \ref{def_CR_dis}) because \small $CR(X_\emptyset)=\sum_{X}[CR(X)P(X)]= \sum_{X}P(X)=1$.\normalsize

\begin{theorem}[Order Independent]
CR is independent of the order of random variables:
\begin{equation*}
CR(X_{a_1};...;X_{a_n})=CR(X_{b_1};...;X_{b_n}),
\end{equation*}
where $[a_1,...,a_n]$ and $[b_1,...,b_n]$ are two different permutations of the sequence $[1,...,n]$.
\end{theorem}

\begin{theorem}[Partition Operation]
\label{thm_partition}
\small
\begin{align*}
&CR(X_1;..;X_k;X_{k+1};..;X_n)\\
&=CR(X_1;..;X_k)CR(X_{k+1};..;X_n)CR(X_1..X_k;X_{k+1}..X_n).
\end{align*}
\normalsize
\end{theorem}
The original $CR(X_1;..;X_k;X_{k+1};..;X_n)$ is partitioned into three parts: (1) the left $CR(X_1;..;X_k)$, (2) the right $CR(X_{k+1};..;X_n)$ and (3) the cut between the left and right $CR(X_1..X_k;X_{k+1}..X_n)$ in which $X_1..X_k$ and $X_{k+1}..X_n$ are two joint variables. This theorem can be used to factorize a graph from top to down.

\vspace{-3mm}
\small
\begin{align*}
&CR(X_1;..;X_k)CR(X_{k+1};..;X_n)CR(X_1..X_k;X_{k+1}..X_n)\\
&=\frac{P(X_1,...,X_k)}{\prod_{i=1}^{k}P(X_i)}\frac{P(X_{k+1},...,X_n)}{\prod_{j=k+1}^{n}P(X_j)}
\frac{P(X_1,..,X_k,X_{k+1},..,X_n)}{P(X_1,..,X_k)P(X_{k+1},..,X_n)}\\
&=\frac{P(X_1,..,X_k,X_{k+1},..,X_n)}{\prod_{l=1}^{n}P(X_l)}=CR(X_1;..;X_k;X_{k+1};..;X_n).
\end{align*}
\normalsize

\vspace*{-3mm}
\begin{theorem}[Merge Operation]
\label{thm_merge}
\small
\begin{align*}
&CR(X_1;..;X_k; X_{k+1};..;X_n)\\
&=CR(X_1;..;X_kX_{k+1};..;X_n)CR(X_k;X_{k+1})
\end{align*}
\normalsize
\end{theorem}
In this theorem two random variables $X_k$ and $X_{k+1}$ are merged into one joint random variable $X_kX_{k+1}$, and a new factor $CR(X_k;X_{k+1})$ is generated. The merge operation can be used to factorize a graph from down to top which is inverse to the Partition Operation. Merging two unconnected nodes implies removing all the conditional independences between them. 

\vspace{-3mm}
\small
\begin{align*}
&CR(X_1;..;X_kX_{k+1};..;X_n)CR(X_k;X_{k+1})\\
&=\frac{P(X_1,.., X_k, X_{k+1},..,X_n)}{P(X_k, X_{k+1})\prod_{i=1}^{k-1}P(X_i)\prod_{j=k+2}^{n}P(X_j)}
\frac{P(X_k, X_{k+1})}{P(X_k)P(X_{k+1})}\\
&=\frac{P(X_1,.., X_k, X_{k+1},..,X_n)}{P(X_1)...P(X_n)}=CR(X_1;..;X_k; X_{k+1};..;X_n).
\end{align*}
\normalsize

\begin{corollary}[Independent Merge] If $X_k$ and $X_{k+1}$ are two independent random variables:
\label{corollary_independent_merge}

\vspace{-3mm}
\small
\begin{align*}
&CR(X_1;..;X_k; X_{k+1};..;X_n)=CR(X_1;..;X_kX_{k+1};..;X_n).
\end{align*}
\normalsize
\end{corollary}
This corollary follows from Merge Operation immediately. As $X_k$ and $X_{k+1}$ are independent, then $CR(X_k;X_{k+1})=1$.

\begin{theorem}[Duplicate Operation]
\label{thm_duplicate}
\small
\begin{align*}
&CR(X_1;..;X_k;..;X_n)=CR(X_1;..;X_k;X_k;..;X_n)P(X_k).
\end{align*}
\end{theorem}
This theorem allows us to duplicate random variables which exist in $CR$. This theorem is useful for manipulating overlapping sub-graphs (Sec. \ref{subsec_ju}).

\vspace{-3mm}
\small
\begin{align*}
&CR(X_1;..;X_k;X_k;..;X_n)P(X_k)=\frac{P(X_1,.., X_k, X_k,..,X_n)}{P(X_k)\prod_{i=1}^{n}P(X_i)}P(X_k)\\
&=\frac{P(X_1,.., X_k, ..,X_n)}{\prod_{i=1}^{n}P(X_i)}=CR(X_1;..;X_k;..;X_n),
\end{align*}
\normalsize
where $P(X_1,.., X_k, X_k,..,X_n)=P(X_1,.., X_k,..,X_n)$ because the logic conjunction operation $\wedge$ is absorptive and we have $(X_k=x_k)\wedge (X_k=x_k)=(X_k=x_k)$.

Here are three Conditional Independence Theorems (CITs) which can be used to reduce the random variables after a partition or merge operation.
\begin{theorem}[Conditional Independence Theorems]
\label{thm_cit}
If $X\ind Y\cond Z$, then the following three Conditional Independence Theorems hold:

\vspace{-5mm}
\small
\begin{align*}
&\textrm{(1)\,\,}CR(X;YZ)=CR(X;Z).\\
&\textrm{(2)\,\,}CR(XY;Z)=CR(X;Z)CR(Y;Z)/CR(X;Y).\\
&\textrm{(3)\,\,}CR(XZ;YZ)=CR(Z;Z)=1/P(Z).
\end{align*}
\normalsize
\end{theorem}

As \small{$X\ind Y\cond Z$}, we have \small{$P(X,Y|Z) = P(X|Z)P(Y|Z)$}. As \small{$P(X,Y|Z) = \frac{P(X,Y,Z)}{P(Z)}$}, \small{$P(X|Z)=\frac{P(X,Z)}{P(Z)}$} and \small{$P(Y|Z)=\frac{P(Y,Z)}{P(Z)}$}, we have \small{$P(X,Y,Z)=P(X,Z)P(Y,Z)/P(Z)$}.
\small
\begin{align*}
&\textrm{(1)\,\,} CR(X;YZ)=\frac{P(X,Y,Z)}{P(X)P(Y,Z)}=\frac{P(X,Z)}{P(X)P(Z)}=CR(X;Z).\\
&\textrm{(2)\,\,} CR(XY;Z)=\frac{P(X,Y,Z)}{P(X,Y)P(Z)}=\frac{P(X,Z)P(Y,Z)}{P(X,Y)P(Z)P(Z)}\\
&=\frac{P(X,Z)P(Y,Z)P(X)P(Y)}{P(X,Y)P(X)P(Z)P(Y)P(Z)}=\frac{CR(X,Z)CR(Y,Z)}{CR(X,Y)}.\\
&\textrm{(3)\,\,} CR(XZ;YZ)=\frac{P(X,Y,Z)}{P(X,Z)P(Y,Z)}\\
&=\frac{1}{P(Z)}=\frac{P(Z,Z)}{P(Z)P(Z)}=CR(Z;Z).
\end{align*}
\normalsize

\begin{theorem}[Unconnected Nodes Theorem]
\label{thm_unt}
Suppose $X_1,X_2$ are two unconnected nodes in $G(X_G,E_G)$, that is there is no direct edge between them; $W,S\subseteq X_G\backslash \{X_1,X_2\}$, where $W\cap S = X_\emptyset$ and $MB\{X_1,X_2\}\subseteq W\cup S$. $MB\{X_1,X_2\}$ is the Markov Blanket of $\{X_1,X_2\}$, then the following identity holds:
\begin{align*}
&CR(Sem\,W; X_1=0; X_2=0; Sem\,S=0)\\
&\times CR(Sem\,W; X_1; X_2; Sem\, S=0)\\
&=CR(Sem\,W; X_1=0; X_2; Sem\,S=0)\\
&\hspace{5mm}\times CR(Sem\,W; X_1; X_2=0; Sem\,S=0),
\end{align*}
where $*=0$ means $*$ is set to an arbitrary but fixed global assignment.
\end{theorem}
\begin{proof}
As $MB\{X_1,X_2\}\subseteq W\cup S$, so $X_1\ind X_2\mathbin{\cond}WS$. \\
For the left side, to each factor we apply partition operation (Thm. \ref{thm_partition}) to split $X_1$ out and then apply the first CIT (Thm. \ref{thm_cit}), then the two original factors on the left side can be factorized as:
\begin{align*}
&CR(Sem\,W; X_2=0; Sem\,S=0)\\
&\times CR(X_1=0;Seq\,S=0\,Seq\,W)\\
&\times CR(Sem\,W;X_2;Sem\,S=0)\\
&\times CR(X_1;Seq\,S=0\,Seq\,W).
\end{align*}
We do the same for the right side. The two original factors on the right side can be factorized as:
\begin{align*}
&CR(Sem\,W;X_2;Sem\,S=0)\\
&\times CR(X_1=0;Seq\,S=0\,Seq\,W)\\
&\times CR(Sem\,W; X_2=0; Sem\,S=0)\\
&\times CR(X_1;Seq\,S=0\,Seq\,W).
\end{align*}
The left side equals the right side.
\end{proof}
This theorem is useful in factorizing Markov Random Fields using co-occurrence rate (Sec. \ref{subsec_mrf}).

Intuitively, conditional probability is an asymmetric concept which matches the asymmetric properties of directed graphs well, while co-occurrence rate is a symmetric concept which matches the symmetric properties of undirected graphs well. Co-occurrence rate also connects probability factorization and graph operations well.
\section{Separate Training}
\label{sec_st}
There are two steps in separate training: (i) factorize the graph using CR-F; (ii) train the factors separately. We use the linear-chain CRFs as the example. 
\begin{figure}[h]
\centering
\includegraphics[scale = 0.40]{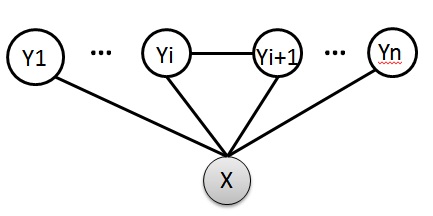}
\caption{Linear-chain CRFs}
\label{fig_crf}
\end{figure}
\normalsize

Linear-chain CRFs can be factorized by CR as follows:
\vspace{-3mm}
\small
\begin{align}
\nonumber
&P(Y_1,Y_2,..., Y_n|X)=CR(Y_1;Y_2;...; Y_n|X)\prod_{j=1}^{n}P(Y_{j}|X)\\
\label{eqn_crf}
&=\prod_{i=1}^{n-1}CR(Y_i;Y_{i+1}|X)\prod_{j=1}^{n}P(Y_{j}|X)
\end{align}
\normalsize
We get the first equation by Def. (\ref{def_conditional_cr}). The second equation is obtained by partition operation (Thm. \ref{thm_partition}) and the first CIT (Thm. \ref{thm_cit}). In practice, we also add the $start$ symbol and $end$ symbol. 

Then we train each factor in a CR factorization separately. We present two training methods. 
\subsection{Exponential Functions}
\label{subsec_ef}
Following \citet{Lafferty:2001:CRF:645530.655813}, factors are parametrized as exponential functions. There are two kinds of factors in Eqn. (\ref{eqn_crf}): the local joint probabilities $P(Y_i;Y_{i+1}|X)$ and the single node probabilities $P(Y_{j}|X)$.

The local joint probabilities can be parametrized as follows:

\vspace{-3mm}
\small
\begin{align*}
\nonumber
&P(Y_i,Y_{i+1}\cond X) = \frac{\exp \sum_{k}\lambda_k f_k(Y_i, Y_{i+1}, X)}{\sum_{Y_iY_{i+1}} \exp \sum_{k}\lambda_k f_k(Y_i,Y_{i+1},X)},
\end{align*}
\normalsize
where $f$ are feature functions defined on $\{Y_i, Y_{i+1}, X\}$, $\lambda$ are parameters and the denominator is the local partition function which covers all the possible pairs of $(Y_i,Y_{i+1})$. In contrast to the global normalization, local partition functions can be reused between training instances whenever the features are the same with respect to $X$.

Similarly, the singleton probabilities is parametrized as follows: 
\vspace{-3mm}
\small
\begin{align*}
\nonumber
&P(Y_i \cond X) = \frac{\exp \sum_{l}\theta_l \phi_l(Y_i, X)}{\sum_{Y_i} \exp \sum_{l}\theta_l \phi_l(Y_i,X)},
\end{align*}
\normalsize
where $\phi$ are feature functions defined on $\{Y_i, X\}$, $\theta$ are parameters and the denominator is the local partition function which covers all the possible tags.

The parameters of each factor are learned separately by following the maximum entropy principle. We use a separate objective function for each factor. This is different from the piecewise training, which learns all parameters by maximizing a single maximum likelihood objective function. To estimate the parameters in $P(Y_i,Y_{i+1}\cond X)$, we maximize the following log objective function:

\vspace{-3mm}
\small
\begin{align}
\nonumber
\mathcal{L}_{sp} =& \sum_{(Y,X)\in D}\sum_{i=1}^{n-1}[\sum_{k}\lambda_k f_k(Y_i, Y_{i+1}, X)\\
\label{eqn_obj}
&-\log \sum_{Y_iY_{i+1}} \exp \sum_{k}\lambda_k f_k(Y_i,Y_{i+1},X)],
\end{align}
\normalsize
where $D$ is the training dataset. As this function is convex, a standard numerical optimization technique, e.g. Limited-memory BFGS, can be applied to achieve the global optimum. The first partial derivative with respect to $\lambda_k$ is given as follows:

\vspace{-3mm}
\small
\begin{align}
\nonumber
&\frac{\partial\mathcal{L}_{sp}}{\partial\lambda_k}=\sum_{(Y,X)\in D}\sum_{i=1}^{n-1}[f_k(Y_i,Y_{i+1}, X) \\
\nonumber
&-\sum_{Y_iY_{i+1}}\frac{\exp \sum_{k}\lambda_k f_k(Y_i,Y_{i+1},X)}{\sum_{Y_iY_{i+1}} \exp \sum_{k}\lambda_k f_k(Y_i,Y_{i+1},X)}f_k(Y_i,Y_{i+1},X)]\\
\label{eqn_der}
&=\sum_{(Y,X)\in D}\sum_{i=1}^{n-1}[f_k(Y_i,Y_{i+1}, X)\\
\nonumber
&-\sum_{Y_iY_{i+1}}P(Y_i, Y_{i+1}|X)f_k(Y_i,Y_{i+1},X)]
=\tilde{E}[f_k]-E_{\Lambda}[f_k]
\end{align}
\normalsize
This derivative is just the difference between the times of occurrences of $f_k$ in the training dataset and the expected times of occurrences of $f_k$ with respect to the estimated distribution $P(Y_i, Y_{i+1}|X)$. We also use Gaussian prior to reduce over-fitting by adding $-\frac{\sum_k \lambda_k^2}{2\sigma^2}$ and $-\frac{\lambda_k}{\sigma^2}$ to Eqn. (\ref{eqn_obj}) and Eqn. (\ref{eqn_der}) respectively. Only the features which have been seen in the training dataset are added to the probability space $P(Y_i,Y_{i+1}|X)$. The parameters in $P(Y_i|X)$ can be learned separately in a similar way. 

\subsection{Fully Empirical}
\label{subsec_fe}
In this training method, we estimate the probabilities in the factors of CR-F by frequencies. Experiment results show that normally this method obtains lower accuracy than the first training method, but this method is very fast (almost instant). This method can be useful for large-scale applications. To estimate $P(Y_i|X)$, if $X$ is observed in the training dataset: $P(Y_i|X)= \frac{\#(Y_i,X)}{\sum_{Y_i}\#(Y_i,X)}$. If $X$ is out of vocabulary (OOV): 
$P(Y_i|X)=\mu_{oov} \frac{\sum_{X'\in A}P(Y_i|X')}{|A|}$, where $A=\{X'; \Phi(Y_i, X')=\Phi(Y_i, X)\}$, $\Phi$ are all the feature functions except the feature function using the word itself.  And to achieve the best accuracy, this method requires an additional parameter $\mu_{oov}$ to adjust the weights between OOV and non-OOV probabilities, where $\mu_{oov}$ is a constant parameter for all OOVs. This parameter can be obtained by maximizing the accuracy on a held-out dataset. In our experiments, $\mu_{oov}$ is between [0.5, 0.65]. Other factors can be learned in a similar way.
\section{Label Bias Problem}
\label{sec_lbp}
One advantage of CRFs over MEMMs is CRFs do not suffer from the label bias problem (LBP)
\citep{Lafferty:2001:CRF:645530.655813}. MEMMs suffer from this problem because they include the factors
$P(Y_{i+1}\cond Y_i, X)$ which are \emph{local conditional probabilities} with respect to $Y$. These local conditional probabilities prefer the $Y_{i}$ with fewer outgoing transitions to others. The extreme case is $Y_{i}$ has only one possible outgoing transition, then its local conditional probability is $1$.  Global normalization as proposed by \citet{Lafferty:2001:CRF:645530.655813} keeps CRFs away from the label bias problem. Co-occurrence Rate Factorization (CR-F) is also unaffected by LBP even though it is a local normalized model. The reason is that, in contrast to MEMMs, the factors in Co-occurrence Rate Factorizations are \emph{local joint probabilities} $P(Y_i,Y_{i+1}|X)$ with respect to $Y$ rather than local conditional probabilities $P(Y_{i+1}\cond Y_i, X)$. This can be seen clearly by replacing the CR factors in Eqn. (\ref{eqn_crf}) with their definitions (Def. \ref{def_CR_dis}):

\vspace{-3mm}
\small
\begin{align}
\nonumber
P(Y|X)&=\prod_{i=1}^{n-1}CR(Y_i;Y_{i+1}|X)\prod_{i=1}^{n}P(Y_{i}|X)\\
\label{eqn_linear_chain_prob}
&=\frac{\prod_{i=1}^{n-1}P(Y_i,Y_{i+1}|X)}{\prod_{j=2}^{n-1}P(Y_j|X)}.
\end{align}
\normalsize
The probabilities of all the transition $(Y_i, Y_{i + 1})$ are normalized in one probability space. That is all the transitions are treated equally. Thus CR-F naturally avoids label bias problem. This is confirmed by experiment results in Sec. (\ref{sec_lbp_exp}). So our method significantly differs from MEMMs in factorization.
\section{Experiments}
\label{sec_exp}
We implement separate training in Java. We also use the L-BFGS algorithm packaged in MALLET \cite{McCallumMALLET} for numerical optimization. CRF++ version 0.57 \cite{crf++} and the piecewise training tool packaged in MALLET are adopted for comparison. All these experiments were performed on a Linux workstation with a single CPU (Intel(R) Xeon(R) CPU E5345, 2.33GHz) and 6G working memory. We denote the first separate training method (Sec. \ref{subsec_ef}) by \textbf{SP2}, the second (Sec. \ref{subsec_fe}) by \textbf{SP1} and the piecewise training by \textbf{PW}.

\subsection{Modeling Label Bias}
\label{sec_lbp_exp}
We test LBP on simulated data following \citet{Lafferty:2001:CRF:645530.655813}. We generate the simulated data as follows. There are five members in the tag space: $\{R1, R2, I, O, B\}$ and four members in the observed symbol space: $\{r, i, o, b\}$. The designated symbol for both $R1$ and $R2$ is $r$, for $I$ it is $i$, for $O$ it is $o$ and for $B$ it is $b$. We generate the paired sequences from two tag sequences: $[R1, I, B]$ and $[R2, O, B]$. Each tag emits the designated symbol with probability of $29/32$ and each of other three symbols with probability $1/32$. For training, we generate 1000 pairs for each tag sequence, so totally the size of training dataset is 2000. For testing, we generate 250 pairs for each tag sequence, so totally the size of testing dataset is 500. As there is no OOVs in this dataset, we do not need a held-out dataset for training $\mu_{oov}$. We run the experiment for 10 rounds and report the average accuracy on tags ($\frac{\#CorrectTags}{\#AllTags}$) in Tab. (\ref{tab:lbp}).

\begin{small}
\begin{table}[h]
\centering
\begin{tabular}{|c|c|c|c|c|}
  \hline
  SP1 & SP2 & CRF++ & PW & MEMMs\\\hline
  95.8\% &95.9\% & 95.9\% & 96.0\% & 66.6\%  \\ \hline
\end{tabular}
\caption{Accuracy For Label Bias Problem}
\label{tab:lbp}
\end{table}
\end{small}
The experiment results show that separate training, piecewise training and the standard training are all unaffected by the label bias problem. But MEMMs suffer from this problem. Here is an example to explain why MEMMs suffer from LBP in this experiment. For an observed sequence $[r, o, b]$, the correct tag sequence should be $[R2, O, B]$. As MEMMs are directed models, they select the first label according to $P(R1|r)$ and $P(R2|r)$. But these two probabilities are almost equal regarding the data generated. So MEMMs may select $R1$ as the first label. Then the next label for MEMMs must be $I$ because: $P(I|R1,o)=1$ and $P(O|R1,o)=0$. That is the second observation $o$ does not affect the result. We can observe the condition $(R1,o)$ in the generated data because $I$ generates $o$ with probability $1/32$. By contrast, separate training based on the co-occurrence rate factorization can make the correct choice because $P(R2,O|r,o)>P(R1,I|r,o)$.

\subsection{POS Tagging Experiment}
We use the Brown Corpus \cite{francis79browncorpus} for POS tagging. We exclude the incomplete sentences which are not ending with a punctuation from our experimental dataset. This results in 34623 sentences. The size of the tag space is 252. Following \citet{Lafferty:2001:CRF:645530.655813}, we introduce parameters for each tag-word pair and tag-tag pair. We also use the same spelling features as those used in \citet{Lafferty:2001:CRF:645530.655813}: whether a token begins with a number or upper case letter, whether it contains a hyphen, and whether it ends in one of the following suffixes: -ing, -ogy, -ed, -s, -ly, -ion, -tion, -ity, -ies. We select 1000 sentences as held out dataset for training $\mu_{oov}$ and fix it for all the experiments of POS tagging. In the first experiment, we use a subset (5000 sentences excluding held-out dataset) of the full corpus (34623 sentences). On this 5000 sentence corpus, we try three splits: 1000-4000 (1000 sentences for training and 4000 sentences for testing), 2500-2500 and 4000-1000. The results are reported in Tab. (\ref{tab:1000-4000}), Tab. (\ref{tab:2500-2500}) and Tab. (\ref{tab:4000-1000}), respectively. In the second experiment, we use the full corpus excluding the held-out dataset and try two splits: 17311-16312 and 32623-1000. The results are reported in Tab. (\ref{tab:17311-16312}) and Tab. (\ref{tab:32623-1000}), respectively.
 
\vspace*{-3mm}
\small
\begin{table}[ht]
\centering
\begin{tabular}{|c|c|c|c|c|}
  \hline
  Metric & SP1 & SP2 & CRF++ & PW\\\hline
  OOVs & 55.9\% & 56.3\%& 39.3\% &47.5\%\\\hline
  non-OOVs & 94.9\% & 94.9\% & 86.8\% &75.3\%\\\hline
  Overall & 86.7\% & 86.8\% & 76.8\% &69.4\%\\\hline
  Time (sec) & 0.4 & 4.3 & 6180.3& 30704.8 \\\hline
\end{tabular}
\caption{1000-4000 Train-Test Split Accuracy}
\label{tab:1000-4000}
\end{table} 

\small
\begin{table}[ht]
\centering
\begin{tabular}{|c|c|c|c|c|}
  \hline
  Metric & SP1 & SP2 & CRF++ & PW\\\hline
  OOVs & 58.2\% & 58.6\%& 43.2\% &49.5\%\\\hline
  non-OOVs & 95.5\% & 95.6\% & 91.2\% &80.0\%\\\hline
  Overall & 90.0\% & 90.2\% & 84.1\% &75.5\%\\\hline
  Time (sec) & 0.6 & 13.25 & 28408.2& 66257.8 \\\hline
\end{tabular}
\caption{2500-2500 Train-Test Split Accuracy}
\label{tab:2500-2500}
\end{table} 
\vspace*{-3mm}
\small
\begin{table}[ht]
\centering
\begin{tabular}{|c|c|c|c|c|}
  \hline
  Metric & SP1 & SP2 & CRF++ & PW\\\hline
  OOVs & 60.5\% & 61.4\%& 44.6\% &52.5\%\\\hline
  non-OOVs & 96.1\% & 96.2\% & 92.9\% &83.0\%\\\hline
  Overall & 91.7\% & 91.9\% & 87.0\% &79.25\%\\\hline
  Time (sec) & 0.95 & 23.5 & 59954.35& 138406.4 \\\hline
\end{tabular}
\caption{4000-1000 Train-Test Split Accuracy}
\label{tab:4000-1000}
\end{table} 

\vspace*{-3mm}
\small
\begin{table}[ht]
\centering
\begin{tabular}{|c|c|c|c|c|}
  \hline
  Metric & SP1 & SP2 & CRF++ &PW\\\hline
  OOVs & 60.8\% & 61.0\%& 62.3\% &50.4\%\\\hline
  non-OOVs & 96.4\% & 96.4\% & 95.3\%&80.8\% \\\hline
  Overall & 94.18\% & 94.2\% & 93.2\%&78.9\% \\\hline
  Time (sec) & 2.2 & 124.6 & 1064384.7&1946706.3 \\\hline
\end{tabular}
\caption{17311-16312 Train-Test Split Accuracy}
\label{tab:17311-16312}
\end{table}

\vspace*{-3mm}
\small
\begin{table}[ht]
\centering
\begin{tabular}{|c|c|c|c|c|}
  \hline
  Metric & SP1 & SP2 & CRF++ &PW\\\hline
  OOVs & 70.1\% & 70.4\%& 71.7\% &59.9\%\\\hline
  non-OOVs & 96.9\% & 96.8\% & 96.1\% &84.0\%\\\hline
  Overall & 95.6\% & 95.6\% & 95.4\%&82.9\% \\\hline
  Time (sec) & 3.9 & 294.9 & 4571806.5 &3791648.2\\\hline
\end{tabular}
\caption{32623-1000 Train-Test Split Accuracy}
\label{tab:32623-1000}
\end{table}
\normalsize

\vspace*{3mm}
The results show that on all experiments, separate training is much faster than the standard training and piecewise training, and achieves better or comparable results.  Tab. (\ref{tab:32623-1000}) shows that with sufficient training data, CRFs performs better on OOVs, but separate training performs slightly better on non-OOVs. As MALLET is in Java and CRF++ is in C++, the time comparison between them is not fair and this is also not the focus of this paper. On the 32623-1000 Split, piecewise training can not converge after more than 300 iterations.

\subsection{Named Entity Recognition}
Named Entity Recognition (NER) also employs a linear-chain structure. In this experiment, we use the the Dutch part of CoNLL-2002 Named Entity Recognition Corpus\footnote{\url{http://www.cnts.ua.ac.be/conll2002/ner/}}. In this dataset, there are three files: ned.train, ned.testa and ned.testb. We use ned.train for training, ned.testa as the held-out dataset for adjusting $\mu_{oov}$ and ned.testb for testing. There are 13221\footnote{Originally, there are 15806 sentences in ned.train. But the piecewise training in MALLET has a bug to decode sentences with only one word. So we have to exclude single word sentences for training and testing.} sentences for training. The size of the tag space is 9. There are 2305 sentences in the held-out dataset and 4211 sentences in the testing dataset. We use the same features as those described in the POS tagging experiment. The results are listed in Tab. (\ref{tab:ner}). 
\begin{table}[H]
\centering
\begin{tabular}{|c|c|c|c|c|}
  \hline
  Metric & SP1 & SP2 & CRF++ & PW         \\\hline
  OOVs & 72.6\% & 72.7\%& 68.3\%& 69.6\%              \\\hline
  non-OOVs & 98.8\% & 98.8\% & 97.0\%& 97.2\%           \\\hline
  Overall & 96.11\% & 96.14\% & 94.1\% & 94.4\%            \\\hline
  Time (sec) & 1.6 & 53.1 & 1070.7 & 4616.5              \\\hline
\end{tabular}
\caption{Named Entity Recognition Accuracy}
\label{tab:ner}
\end{table}
\normalsize
On the NER task, separate training is the fastest and obtains best results. Piecewise training obtains slightly better result than the standard training method which is consistent with the results reported by \citet{Sutton05piecewisetraining}.

%
%

\section{Relationship To Other Factorization Methods}
\label{sec_re}
Since a co-occurrence rate relation includes any statement one can make about independence relations, it is not a surprise that we can rework other factorization methods, such as Junction Tree Factorization and Markov Random Fields, using it. In this section, we sketch how to obtain factors in Junction Tree and MRFs using the operations of CR-F.
%
%
%
\subsection{CR-F and Junction Tree}
\label{subsec_ju}
Suppose we constructed a junction tree $X_G$ which satisfies the \emph{running intersection property} \cite{rip}, that is, there exists a sequence $[C_1,C_2,...,C_n]$, where $C_1,C_2,...,C_n$ are all maximal cliques in $X_G$, and if we separate $C_i$ out from $X_G$ in the order of this sequence, there exists a clique $C_x$, where $i<x\leq n$, and separator nodes $S_i=C_i\cap C_x$, satisfying $(C_i\backslash S_i)\cap C_j=\emptyset$ for all $C_j, i<j\leq n$. We can factorize $X_G$ using CR-F as follows:

\textbf{Step 0}: $P(X_G)=CR(Sem\, X_G)\prod_{X\in X_G}P(X)$.

\textbf{Step 1}: For $i=1$ to $n-1$, duplicate the separator nodes $S_i$ (Thm. \ref{thm_duplicate}):

\vspace{-3mm}
\small
\begin{align*}
\label{eqn_junction_tree_dup}
&CR(Sem\,\cup_{j=i}^{n} C_j)=CR(Sem\,S_i;Sem\,\cup_{j=i}^{n} C_j)\prod_{X\in S_i}P(X),
\end{align*}
\normalsize
and partition $C_i$ out (Thm. \ref{thm_partition}): 

\vspace{-3mm}
\small
\begin{align*}
&CR(Sem\,S_i;Sem\,\cup_{j=i}^{n} C_j)\\
&=CR(Sem\,\cup_{j=i+1}^{n} C_j)CR(Sem\,C_i)(Seq\,C_i;Seq\,\cup_{j=i+1}^{n} C_j)\\
&=CR(Sem\,\cup_{j=i+1}^{n} C_j)CR(Sem\,C_i)CR(Seq\,S_i;Seq\,S_i)\\
&=CR(Sem\,\cup_{j=i+1}^{n} C_j)CR(Sem\,C_i)\frac{1}{P(S_i)}.
\end{align*}
\normalsize
We obtain the second equation by Thm. (\ref{thm_cit}) as $S_i$ completely separate $C_i$ from the remaining part of the graph ($\cup_{j=i+1}^{n} C_j$). The running intersection property guarantees there exist separator nodes $S_i$ for each $C_i$.

Finally, we can get the factors on junction tree cliques. For $C_i\neq C_n$ and $C_n$:

\vspace{-3mm}
\small
\begin{align*}
&\phi_{C_i}(C_i)=CR(Sem\,C_i)\frac{\prod_{X\in C_i}P(X)}{P(S_i)}=\frac{P(C_i)}{P(S_i)},\\
&\phi_{C_n}(C_n)=CR(Sem\,C_n)\prod_{X\in C_n}P(X)=P(C_n).
\end{align*} 
\normalsize
Thus the joint probability can be written as \small$P(X_G)=\frac{\prod_{i=1}^{n}P(C_i)}{\prod_{j=1}^{n-1}P(S_j)}$\normalsize , where $C_i$ is a maximum clique and $S_j$ is a separator. This result is similar to that obtained by Shafer-Shenoy propagation except that the factors obtained by CR-F are local joint probabilities rather than just positive functions. These local joint probabilities can be normalized locally and trained separately.

\subsection{CR-F and MRF}
\label{subsec_mrf}
The joint probability over Markov Random Fields can be written as products of positive functions over maximal cliques:
\begin{equation*}
P(X_G)= \frac{1}{Z}\prod_{mc\in MC}\phi_{mc}(mc),
\end{equation*}
where $MC$ are all the maximal cliques in $G$ including $X_{\emptyset}$, $\phi_{mc}(mc)$ is a positive potential defined on $mc$, and $Z$ is the partition function for normalization.

This factorization can be obtained by CR-F as follows: 

Firstly, the following identity holds obviously:

\vspace{-3mm}
\small
\begin{equation}
\label{eqn_m_factors}
1=\prod_{S\in \mathcal{P}(X_G)\backslash X_G}(\frac{CR(Sem\,S; Sem\, X_G\backslash S=0)}{CR(Sem\,S; Sem\,X_G\backslash S=0)})^U, 
\end{equation}
\normalsize
where $U={2^{\mathbin{|}X_G\mathbin{|}-\mathbin{|}S\mathbin{|}-1}}$. The right side of this identity is denoted by $M$, then:

\vspace{-3mm}
\small
\begin{equation}
\label{eqn_factors}
P(X_G)=M\times CR(Sem\,X_G)\times \prod_{i=1}^{|X_G|} P(X_i).
\end{equation}
\normalsize
Then we group these factors into proper scopes. The proper scopes are $\mathcal{P}(X_G)$. For each scope $sc\in \mathcal{P}(X_G)$, we group the following factors in Eqn. (\ref{eqn_factors}) into $sc$:
\begin{equation*}
\{CR(Sem\,S; Sem\, X_G\backslash S=0)^{(-1)^{|sc|-|S|}}, S\in \mathcal{P}(sc)\}.
\end{equation*}
The following two binomial identities guarantee that all the factors are just be grouped into scopes $\mathcal{P}(X_G)$:

\vspace{-3mm}
\small
\begin{align*}
&2^N=(1+1)^N=\binom{N}{0}+\binom{N}{1}+...+\binom{N}{N},
\end{align*}
\normalsize

\vspace{-3mm}
\small
\begin{align*}
&0^N=(1-1)^N=\binom{N}{0}-\binom{N}{1}+...+(-1)^N \binom{N}{N}.
\end{align*}
\normalsize
where $N=\mathbin{|}X_G\mathbin{|}-\mathbin{|}S\mathbin{|}$. We go on to prove if $sc$ is not a clique, then all the factors grouped into $sc$ cancel themselves out. If $sc$ is not a clique, there must exist two unconnected nodes $X_a$ and $X_b$ in $sc$. Let $W\in \mathcal{P}(sc\backslash \{X_a,X_b\})$, then all the factors selected into $sc$ can be categorized into four types: $W$, $W\cup \{X_a\}$, $W\cup \{X_b\}$ and $W\cup\{X_a,X_b\}$, and they can be written as follows:

\vspace{-3mm}
\small
\begin{align*}
&\prod_{sc\in NC}\prod_{S\in \mathcal{P}(sc)}CR(Sem\,S; Sem\, X_G\backslash S=0)^{(-1)^{|sc|-|S|}} \\
&=\prod_{sc\in NC}\prod_{W\in J}(\frac{CR(Sem\,W;X_a=0;X_b=0; Sem\,X^*=0)}{CR(Sem\, W;X_a=0;X_b;Sem\,X^*=0)}\\
&\hspace{20mm}\frac{CR(Sem\,W;X_a;X_b;Sem\, X^*=0)}{CR(Sem\,W;X_a;X_b=0;Sem\,X^*=0)})^{-1^*},
\end{align*}
\normalsize
where $J=\mathcal{P}(sc\backslash \{X_a,X_b\})$, $NC$ are all the non-clique scopes in $\mathcal{P}(X_G)$, and $X^*=X_G\backslash (W\cup  \{X_a,X_b\})$. $X=0$ means $X$ is set to an arbitrary but fixed assignment. This assignment is global and called global configuration.  Only the relative positions of the four factors are important, thus we use $-1^*$ to represent the power. According to Thm. (\ref{thm_unt}), this equation equals $1$, so the factors in non-clique scopes cancel themselves out. Now only the factors selected into clique scopes are left, which can be further grouped into maximum cliques.

Since the factors obtained in MRFs depends on a global fixed configuration, these factors are not really independent and thus can not be trained separately.

\section{Conclusions}
\label{sec_co}
In this paper, we proposed the novel Co-occurrence Rate Factorization (CR-F) for factorizing undirected graphs. 
Based on CR-F we presented the separate training for scaling CRFs. Experiments show that separate training (i) is unaffected by the label bias problem, (ii) speeds up the training radically and (iii) achieves competitive results to the standard and piecewise training on linear-chain graphs. We also obtained the factors in MRFs and Junction Tree using CR-F. This shows CR-F can be a general framework for factorizing undirected graphs.

\section{Future Work}
\label{sec_fu}
In this paper, we present separate training on linear-chain graphs. Separate training can be easily extended to tree-structured graphs. In the future, we will generalize separate training to loopy graphs. Briefly, using
Thm. (\ref{thm_cond_op}), we can break loops. When a node in a loop is partitioned out, we need to bring it back as a condition to avoid adding a new edge. In this way we can keep the factorization exact.

\nocite{langley00}

\bibliography{CR}
\bibliographystyle{icml2013}

\end{document}